\documentclass[11pt,a4paper]{article}
\usepackage[hyperref]{emnlp-ijcnlp-2019}
\usepackage{times}
\usepackage{latexsym}
\usepackage{amsmath}
\usepackage{url}
\usepackage{mathdots}
\usepackage{yhmath}
\usepackage{cancel}
\usepackage{color}
\usepackage{siunitx}
\usepackage{array}
\usepackage{multirow}
\usepackage{amssymb}
\usepackage{gensymb}
\usepackage{tabularx}
\usepackage{booktabs}
\usepackage{svg}

\aclfinalcopy

\title{Improving Neural Story Generation by\\ Targeted Common Sense Grounding}

\author{Huanru Henry Mao, Bodhisattwa Prasad Majumder, \\\textbf{Julian McAuley, Garrison W. Cottrell} \\
  Department of Computer Science and Engineering \\
  UC San Diego \\
  {\tt \{hhmao, bmajumde, jmcauley, gary\}@eng.ucsd.edu}}

\date{}

\begin{document}
\maketitle
\begin{abstract}
    Stories generated with neural language models have shown promise in grammatical and stylistic consistency.
    However, the generated stories are still lacking in \emph{common sense reasoning}, e.g.,~they often contain sentences deprived of world knowledge.
    We propose a simple multi-task learning scheme to achieve quantitatively better common sense reasoning in language models by leveraging auxiliary training signals from datasets designed to provide common sense grounding.
    When combined with our two-stage fine-tuning pipeline, our method achieves improved common sense reasoning and state-of-the-art perplexity on the \emph{WritingPrompts} \cite{DBLP:conf/acl/LewisDF18} story generation dataset.
\end{abstract}

\section{Introduction}
\emph{Story generation} is the task of automatically producing compelling creative writing.
Recent advances in language modeling have yielded thematic and stylistic coherence in story generation through large scale pretraining of Transformer models \cite{DBLP:journals/corr/VaswaniSPUJGKP17}.
The recent introduction of the General Pre-trained Transformer v2 (GPT2) \cite{gpt2}---a high-capacity Transformer trained on a large, diverse corpus of text crawled from the web (called WebText)---is capable of generating stylistically coherent text but commonly produces text with logical inconsistencies.
For example, in one sample the model writes: ``\textit{It was a sunny, warm summer \underline{night}}". 
Obviously, this writing is nonsense as it cannot be sunny at night.
The lack of common sense reasoning in such a strong language model suggests that minimizing next-token perplexity alone may be insufficient in producing models that can compose sensible stories.

\begin{figure}[t!]
    \centering
    \includegraphics[width=\columnwidth]{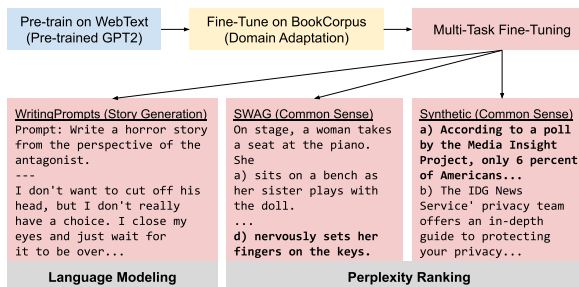}
    \caption{Our two-stage fine-tuning pipeline with auxiliary multi-task learning. For perplexity ranking examples, bolded text indicates the correct answer. For \textbf{synthetic}, the correct choice is written by a human, and the wrong choice is generated by a neural network.}
\label{fig:tasks}
\end{figure}

In this paper, we consider the challenge of \textit{common sense reasoning} (CSR) in language modeling for story generation. 
Unlike other work in the CSR literature \cite{DBLP:journals/corr/abs-1904-01172}, which evaluates CSR in isolation, we are specifically interested in a generative model's likelihood of producing text that exhibits common sense.
Evaluating common sense qualitatively in a model's samples is difficult, as it is subject to human bias and dependent on sampling procedure.
We propose evaluating the common sense of a model automatically by ranking the model's perplexity on spurious (plausible, but nonsense) text completions from SWAG \cite{DBLP:journals/corr/abs-1808-05326} and Story Cloze \cite{DBLP:journals/corr/MostafazadehCHP16} datasets, which are designed for common sense grounding.

\textbf{Our contributions} are as follows:
We propose a simple way to define \textit{better} CSR in generative models, which leads to an auxiliary multi-task objective to directly bias our model to generate text with better common sense.
When fine-tuning is combined with multi-task learning in a two-stage pipeline, we improve the model's CSR and outperform state-of-the-art perplexity on the WritingPrompts \cite{DBLP:conf/acl/LewisDF18} dataset.\footnote{Our source code is at \url{https://github.com/calclavia/story-generation}.}

\section{Tasks}
\subsection{Primary Task: Language Modeling}
Our primary task is to perform language modeling \cite{Elman_90, DBLP:journals/jmlr/BengioDVJ03, DBLP:journals/corr/DaiL15a} on the WritingPrompts dataset.
A language model learns to assign the probability of a text sequence \(X = {x_1,\ldots,x_T}\) using the conditional probability factorization:
\begin{equation}
\label{eq:lm_prob}
P(X) = \prod_{t=1}^{T}{P(x_{t} |x_{1:t-1})}.
\end{equation}
We train our model using a standard cross-entropy loss between next-step true tokens and predicted probabilities given current tokens.

WritingPrompts \cite{DBLP:conf/acl/LewisDF18} is a dataset of prompts and short stories crawled from Reddit.
The aim of the dataset is to produce a story given a free-text prompt.
We reduce this conditional text generation task into a generic language modeling task by simply concatenating the prompt before the story and treating a prompt-story pair as one input to the Transformer decoder model. This human-readable format (example in Figure \ref{fig:tasks}) is chosen because GPT2 may have been trained on similarly formatted text from the web. When sampling, we can either seed the model with a prompt or allow it to generate its own prompt.

\subsection{Auxiliary Task: Perplexity Ranking}
Our auxiliary task aims to bias the model to produce text with better common sense (which we refer to as \textit{sensible} text).
Given a set of text sequences \(S = \{S_1,\ldots,S_N\}\), where \(S_1\) is a sensible text sequence and the rest \(S_2,\ldots,S_N\) are spurious text sequences, we operationally define \textit{better} as the model assigning higher probability (Eq.~\ref{eq:lm_prob}) to $S_1$ versus the average probability over spurious text sequences $S_2,\ldots,S_N$.
Using this definition, it is possible to directly optimize the model to assign \(P(S_1)\) to be higher than any \(P(S_i)\).
Formally, we define the probability of the model choosing the correct sequence \(S_1\) over spurious sequences as the softmax over the length-normalized log probabilities of all plausible sequences:
\begin{equation}
\label{eq:aux_ppl_rank}
\frac{\exp ({\frac{1}{T_1} \log P(S_1)}) }{\sum^N_{i=1}{\exp(\frac{1}{T_i} {\log P(S_i)}})}
\end{equation}
where $T_i$ refers to the length of the text sequence \(S_i\). Thus, we can practically minimize the negative log-likelihood of Eq. \ref{eq:aux_ppl_rank} by reusing the same softmax layer used for the primary language modeling task. We refer to this objective as \textit{perplexity ranking} as it constrains the model to rank sensible text to have lower perplexity than spurious ones.

\smallskip
\noindent
\textbf{SWAG: }
In order to train on this auxiliary objective, we need training examples in the format of multiple choice questions, where the correct choice corresponds to the text with the best common sense.
We choose the SWAG dataset \cite{DBLP:journals/corr/abs-1808-05326}, a dataset that provides 4-way multiple choice common sense questions that are adversarially filtered to seem plausible to language models.
Unlike other CSR datasets \cite{DBLP:journals/corr/abs-1811-00937}, SWAG forms its question and answer as two full sentences, which we can concatenate into a single string to find its probability. This makes it suitable for perplexity ranking.

\smallskip
\noindent
\textbf{Synthetic: }
To facilitate perplexity ranking on SWAG, we additionally use a synthetic dataset that consists of 250K pairs of human written text from WebText and samples generated by the original 1.5B parameter version of the GPT2 model.\footnote{We obtained the samples from \url{https://github.com/openai/gpt-2-output-dataset}}
These samples are many paragraphs long and truncated to a maximum of 1024 tokens.
We frame these pairs as a 2-way classification problem and train the model by perplexity ranking to assign higher likelihood to human written text over synthetic examples.
The assumption we make is that human written text is more sensible than text written by neural language models.
Hence, on average, this promotes the model to assign higher probabilities to sensible text.

\begin{table}[t!]
\small
\centering
\begin{tabularx}{\columnwidth}{Xrr}
\toprule
\bf Dataset     & \bf Size          & \bf Role      \\ \midrule
WritingPrompts  & 272K Stories      & Story Generation   \\
BookCorpus      & 10K Books         & Domain Adaptation      \\
SWAG            & 73K Questions     & Common Sense       \\
Synthetic       & 250K Pairs        & Common Sense   \\
\bottomrule
\end{tabularx}
\caption{Datasets, training set sizes and their roles}
\label{tab:datatsets}
\end{table}

\section{Training Pipeline}
We introduce a two-stage training pipeline (Figure \ref{fig:tasks}) to improve model performance both in terms of perplexity and CSR on story generation.
Our pipeline uses four different datasets (Table \ref{tab:datatsets}), each of which plays a role in improving model performance.
Our model architecture is the 117M parameter version of GPT2, using the pre-trained weights provided by \citet{gpt2}. We refer readers to \citet{DBLP:journals/corr/VaswaniSPUJGKP17} for details of the Transformer architecture.

We first perform intermediate fine-tuning \cite{DBLP:journals/corr/abs-1811-01088,DBLP:journals/corr/abs-1801-06146} of the pre-trained GPT2 on BookCorpus \cite{DBLP:conf/iccv/ZhuKZSUTF15} as a method of domain adaptation from WebText to the domain of stories. BookCorpus is a dataset that contains over 10,000 free books crawled from the web.\footnote{We crawled BookCorpus from \url{https://www.smashwords.com/}} We train on this corpus using our language modeling objective.
Next, we fine-tune on the target WritingPrompts dataset with a multi-task learning objective.
We alternate training between the language modeling objective on WritingPrompts and perplexity ranking on SWAG and our synthetic dataset.
Training details and hyperparameters are in the appendix.

\section{Evaluation}
We perform three types of evaluation on the model to assess its readability, reliance on the prompt (prompt ranking) and CSR.

Readability is measured in terms of model perplexity on the test set of WritingPrompts.
Because GPT2 uses subword tokenization \cite{DBLP:conf/acl/SennrichHB16a}, it is not directly comparable to the word-level perplexity obtained in \citet{DBLP:conf/acl/LewisDF18}. We estimate the corresponding word-level perplexity by taking the product of each subword's probabilities to obtain probabilities for each word. Both sub-word perplexity 
%(\textbf{WP SW PPL}) 
and word-level perplexities
%(\textbf{WP Word PPL}) 
are reported in our experiments.

Prompt ranking \cite{DBLP:conf/acl/LewisDF18} assesses how well a model matches a story to its given prompt. This is measured by computing the likelihood of stories conditioned under ten different prompts, nine of which are randomly sampled and one is the true prompt. Following \citet{DBLP:conf/acl/LewisDF18}, we count a random story sample as correct when it ranks the true prompt with the lowest perplexity. We compute the accuracy from 1000 random samples.

CSR is evaluated on two multiple choice datasets -- SWAG and Story Cloze \cite{DBLP:journals/corr/MostafazadehCHP16}. We rank the perplexity computed by the model for each example and count it as correct if the lowest perplexity matches the answer. The SWAG validation set provides a proxy of how well the model generalizes in CSR to unseen examples. To ensure generalization beyond SWAG, we also perform \textit{zero-shot} evaluation on the Winter 2018 Story Cloze validation set. Story Cloze consists of 5-sentence stories with correct and spurious endings. It is similar to SWAG but serves as an in-domain dataset to specifically test the model's performance at CSR in story telling.

\section{Results and Discussion}
\begin{table}[t!]
\begin{center}
\small
\begin{tabularx}{\columnwidth}{X}
\toprule
\textbf{Premise: }\textit{John and Billy became very skilled at beer pong. They entered a contest in college. They won the contest and advanced to the next level. The next level sent them to Vegas.}\\
\midrule
\textbf{GPT2 $\rightarrow$ BC $\rightarrow$ WP output:}

1.~They would fall.

2.~Later, they figured out what it was all about.
\smallskip\\
\textbf{GPT2 $\rightarrow$ BC $\rightarrow$ WP + SWAG + SYNTH output:}

1.~They have been ranked number one in their respective leagues and are considered the best in their respective countries.

2.~They then moved to the bars.\\
\bottomrule
\end{tabularx}
\end{center}
\caption{\label{tab:samples} Top two highest likelihood story completions from 10 random completion samples generated by our models when primed with a premise from the Story Cloze validation set.}
\end{table}

\begin{table*}[t!]
\begin{center}
\small
\begin{tabularx}{\textwidth}{Xccccc}
\toprule
\bf Models & \bf SW PPL & \bf Word PPL & \bf Prompt Ranking & \bf SWAG & \bf Story Cloze \\
\midrule
Fusion Model \cite{DBLP:conf/acl/LewisDF18} & - & 36.6 & 16.3\% & - & - \\ 
 \midrule

GPT2 & 35.57 & 51.29* & 49.8\% & 48.1\% & 58.8\% \\ 
GPT2 $\rightarrow$ BC & 29.10 & 42.01* & 62.7\% & 50.5\% & 59.6\% \\ 
GPT2 $\rightarrow$ WP & 21.68 & 30.65* & 80.0\% & 49.8\% & 58.3\%\\ 
GPT2 $\rightarrow$ BC $\rightarrow$ WP & 20.79 & 29.56* & \textbf{80.6\%} & 51.4\% & 59.1\% \\ 
GPT2 $\rightarrow$ BC $\rightarrow$ WP + SWAG & \textbf{20.78} & \textbf{29.52*} & 78.9\% & 75.3\% & 63.2\% \\ 
GPT2 $\rightarrow$ BC $\rightarrow$ WP + SWAG + SYNTH & \textbf{20.78} & 29.63* & 80.1\% & \textbf{76.3\%} & \textbf{64.1\%} \\ 
\bottomrule
\end{tabularx}
\end{center}
\caption{\label{tab:results} Performance of models on the test set of WritingPrompts and validation set of SWAG and Story Cloze. \textbf{SW PPL} and \textbf{Word PPL} refer to sub-word and word-level perplexity on WritingPrompts respectively. WP refers to WritingPrompts, BC refers to BookCorpus and SYNTH refers to training with our 250K synthetic examples. The asterisk \textit{*} refers to an estimated score derived from BPE PPL.}
\end{table*}

We analyze our pipeline and report the results in Table \ref{tab:results}. We also generate stories by sampling from our model using nucleus sampling with \(p=0.9\) \cite{DBLP:journals/corr/abs-1904-09751}. We present example story completions in Table \ref{tab:samples} and full sampled stories in our appendix and Table \ref{tab:full_sample1}.

\noindent
\textbf{Pre-training:} 
We compare our models with the attention-based Fusion Model \cite{DBLP:conf/acl/LewisDF18}, which has been designed for and trained on WritingPrompts. We observe that a pre-trained GPT2 performing zero-shot inference on WritingPrompts (GPT2 in Table~\ref{tab:results}) is a strong baseline.
By fine-tuning GPT2 on WritingPrompts (GPT2 $\rightarrow$ WP), we outperform the Fusion Model in perplexity.
All our models outperform the Fusion Model in prompt ranking, which suggests that task-specific models are unnecessary given pre-training.

\noindent
\textbf{Intermediate Fine-Tuning:}
The first stage in our pipeline performs intermediate fine-tuning of GPT2 on BookCorpus (GPT2 $\rightarrow$ BC in Table~\ref{tab:results}). To confirm that intermediate fine-tuning helps downstream performance, we evaluate the zero-shot performance of the model on WritingPrompts.
This yields perplexity and prompt ranking improvements compared to GPT2, demonstrating successful domain adaptation.
Performing two-stage fine tuning (GPT2 $\rightarrow$ BC $\rightarrow$ WP) further improves perplexity and CSR. We hypothesize the improvement in CSR is due to BookCorpus being a higher quality dataset written by authors when compared against WebText.

\noindent
\textbf{Multi-tasking Fine-Tuning:}
Performing multi-task learning on WritingPrompts and SWAG (GPT2 $\rightarrow$ BC $\rightarrow$ WP + SWAG) unsurprisingly yields significant improvements on the SWAG validation set. More importantly, the zero-shot performance on Story Cloze also improved, indicating that it was able to generalize its common sense knowledge. We also see qualitatively improved results when generating story completions (Table \ref{tab:samples}). The addition of the synthetic dataset we introduced (GPT2 $\rightarrow$ BC $\rightarrow$ WP + SWAG + SYNTH) further boosts performance on CSR. Other metrics are negligibly affected by the auxiliary tasks.

\section{Related Work}
\noindent
\textbf{Story Generation:}
Recent work in neural story generation \cite{DBLP:journals/corr/KirosZSZTUF15,DBLP:conf/aaai/Roemmele16} has shown success in using hierarchical methods \cite{DBLP:journals/corr/abs-1811-05701, DBLP:conf/acl/LewisDF18} to generate stories. In these schemes, a neural architecture is engineered to first generate an outline or a prompt, then to expand the prompt into a full-length story. Our work performs hierarchical generation, but our main focus is on achieving better common sense in the generated text rather than engineering task-specific architectures.

\noindent
\textbf{Common Sense Reasoning:}
Common sense reasoning (CSR) has been studied through many benchmarks such as SWAG \cite{DBLP:journals/corr/abs-1808-05326}, Story Cloze \cite{DBLP:journals/corr/MostafazadehCHP16}, the Winograd Schema Challenge \cite{levesque2012winograd}, and CommonsenseQA \cite{DBLP:journals/corr/abs-1811-00937}. 
Recent methods \cite{DBLP:journals/corr/abs-1802-05365,gpt1} on these benchmarks focus on large-scale pre-training of language models. They show that transfer learning is an effective means to improve CSR and our fine-tuning pipeline builds upon these techniques. 
Our results on SWAG and Story Cloze are far from state-of-the-art \cite{DBLP:journals/corr/abs-1810-04805}.
However, our aim is not to directly tackle SWAG or Story Cloze, but instead to use it as a constraint on our model and a proxy to estimate the likelihood of generating sensible text.

\noindent
\textbf{Multi-task Learning:}
Multi-task learning (MTL) introduces inductive bias in a model, helps reduce overfitting and increases robustness \cite{DBLP:conf/icml/Caruana93, DBLP:journals/corr/Ruder17a, mccann2018natural}.
Our work builds upon MTL principles as we introduce auxiliary tasks to specifically tackle CSR \cite{bingel2017identifying}. 
Contrary to conventional auxiliary multi-task learning \cite{DBLP:journals/corr/abs-1801-06146}, which adds an additional classifier to the final representations of the model for learning auxiliary tasks, we use perplexity ranking, which does not introduce additional parameters. We argue that perplexity ranking, by definition, guarantees that the model will generate sensible text with a higher probability.

\section{Conclusion}
We demonstrated a simple way to improve the likelihood that a particular language model generates sensible text.
We introduced a training pipeline that leverages transfer and multi-task learning to improve perplexity and CSR.
One limitation of our approach is its dependence on SWAG, which only provides short-term common sense grounding.
While our synthetic dataset may mitigate this issue, story-level coherence is unlikely to be achieved using this method alone.
We leave this investigation for future work.

\begin{table*}[t]
\begin{center}
\begin{tabularx}{\textwidth}{X}
\toprule
\textbf{Prompt:} Aliens start abducting humans.
\\
\midrule
Steve smashed his Xbox controller at the table. He lit a smoke and took another hit. Two exes were in the room, just outside the door.

``What are you doing?" he said. ``Get back in there."

``I-I've never actually seen any of the aliens," he replied, dispelling the smoke from his mouth.

The two aliens were smirking. ``Is this your first time?" the ex asked.

``The last one tried to kill us."

The aliens looked at each other. ``Are you sure you are ok? It looks like you are severely hurt."

``Yeah, I'm fine. I just saw the door open. I'm just fine," he replied.

``We're fine? You do that every day, Steve, but we still haven't gotten past this door," the ex said, resting a hand on the
UFO that stood in the doorway.

``I'm fine."

``Good! I've never seen anything like it, have I Steve?"

Steve stood up. ``Yes I have." He picked up his controller.

``Ok, shoot!" the aliens yelled. ``This is no time to hesitate."

Steve put the controller to his ear. ``Who are you? What are you doing here?"

The aliens pulled the gun towards him. "We-we're here to kidnap you."

Steve put his hands on the UFO, and pulled the trigger. The gun went off. The aliens, scared, immediately realized where they were and ran towards the other two aliens. They kept running. Steve continued on his way towards his childhood toy house. When he was just a teen, he had built his own helicopter. He always felt they had lost. The abandoned building on the edge of the highway he and the ex had entered had been the perfect landing spot. He didn't know where they were. He didn't know how they were going to get out of there. There were bodies everywhere.
\\
\bottomrule
\end{tabularx}
\end{center}
\caption{Sample generated by GPT2 $\rightarrow$ BC $\rightarrow$ WP + SWAG + SYNTH primed with the same prompt as \citet{DBLP:conf/acl/LewisDF18}.}
\label{tab:full_sample1}
\end{table*}

% Output samples in appendix

\noindent\textbf{Acknowledgements.} This work is partly supported by NSF \#1750063, the Guangzhou Science and Technology Planning Project \#201704030051, NSF awards CNS-1730158, ACI-1540112,  ACI-1541349, OAC-1826967, the University of California Office of the President, and the University of California San Diego's California Institute for Telecommunications and Information Technology/Qualcomm Institute. We thank CENIC for the 100Gpbs networks and all the reviewers for their constructive suggestions.

\bibliography{emnlp-ijcnlp-2019}
\bibliographystyle{acl_natbib}

\section{Training Details}
In this section, we describe our training method in detail. For all experiments, we use a batch size of 16 and train on 8 NVIDIA 2080Ti GPUs.

Our training pipeline consists of two stages. In the first stage, we perform domain adaptation using BookCorpus \cite{DBLP:conf/iccv/ZhuKZSUTF15} dataset. We fine-tune following techniques from \citet{DBLP:journals/corr/abs-1801-06146}. We use a triangular learning rate with 1,000 iterations of warm-up to a maximum learning rate of \(5 \times 10^-5\) then decay the learning rate to zero for the next 99,000 training iterations.

In the second stage, we perform multi-task fine-tuning on our primary task (WritingPrompts) and auxiliary tasks (SWAG and our synthetic dataset).
We alternate training between these three datasets.
We train using our language modeling objective on WritingPrompts every iteration. 
For every 15 iterations, we train on the synthetic dataset and every 30 iterations we train on the SWAG training set. In effect, this places more importance on the primary task over the auxiliary tasks.
We use the same learning rate schedule as in the first stage and apply early stopping using the validation set of WritingPrompts.

\section{Samples}
We include full stories generated by two of our models with and without the auxiliary multi-task objectives. These samples are unconditional, e.g.~the model also generates the prompts. For comparison with \citet{DBLP:conf/acl/LewisDF18}, we also generate samples from our best model with the exact same prompt used to seed their samples.

\begin{table*}
\begin{center}
\begin{tabularx}{\textwidth}{X}
\toprule
\textbf{Prompt:} Today, you are a teenager living in a nation that's population is estimated at close to 1 billion. You
are also in the midst of one of the most brutal wars in history, with almost no sign of stopping until a nuke strikes the United States.
\\
\midrule
Most people didn't realize the situation at all. It was like they were surrounded. Or maybe that was me. Probably another setup.

The third one, the decision was me, that'd been a week ago. A very long time ago, I was in the United States and would have died back there. I had no idea what was going on but I had gotten fed up with them. Well, I was glad I had left. They hadn't gotten fed up with me. They all think they can keep me to the gates.

"Stay with us!" I shouted to the crowds in the crowd. As many of them watched as I marched through the crowds, those I could only see as they walked toward me, would come closer to see me. I had children on my shoulder now.

I stepped into the classroom to report for duty. ``First reports have been very rough on my family. They all broke in and assaulted one of the teachers during their lunch break. That was the beginning of it, right? So they can't trust us to stay with them and listen to them, right?" The classroom was covered in bullet holes and broken desks. Everyone from the school to
the Pentagon had come to help.

``So who are they? I mean we sent them into Washington and Washington D.C. and they've taken over our city and our houses. I don't know, that doesn't sound good. We're going to the truck to get to and get them out of there. Even if they try to run
out we're going to get them back out. If they try anything else they'll be shot, if they try to escape or even worse. I hear there's a couple of US soldiers out there looking for me. I'll just tell them to send some valuables to the embassy. They have better phones. What if they try to make it to the bus station or what?"

The people in the classroom didn't take this any more seriously than I had. Their survival and technology would be long over the horizon. It was a race that only the defense budget could have prepared for. "What's the worse that could happen? " I
asked one of the doctors. " I can't risk it. " The doctor had stopped talking to me. " Do you know where we are? Do you know how long it's been?"

The state almost overwhelmed the building like I had. I knew that far outside the walls, on the other side of the nation, there were more. But it's a safe bet that what I was seeing now was some kinda, ``Plenty of people" army, occupying their homes and running them off. They couldn't get rid of the CIA or, or, no, the world government, it would have been too easy. Their military didn't even have such potential, no army could have. It would have been too easy.

They weren't people. No, there were hundreds of people that would have been so easily overcome.

"We have to go down to the country base. The embassies would be taken care of."

I started heading down to the bus station.
\\
\midrule
\textbf{Prompt:} Write a short scene in which one character reduces another to uncontrollable sobs without touching him
or speaking.
\\
\midrule
It was the first day of finals, and his class had stopped for the night. But that's ok; it didn't feel like his class mattered, it just seemed to wind up passing without him noticing. He would write away and miss his chance at one of his halloween
parties, but he found himself crushed by the feeling of his class being forgotten.

It wasn't an act of love, or a lust, or an angst; it was just all that remained of it all. He smiled at the young woman in
his class and said the first thing he thought of before going off to dinner.

"You're such a brat."\\
\bottomrule
\end{tabularx}
\end{center}
\caption{Samples generated by GPT2 $\rightarrow$ BC $\rightarrow$ WP.}
\end{table*}

\begin{table*}
\begin{center}
\begin{tabularx}{\textwidth}{X}
\toprule

\textbf{Prompt:} Someone can make you be who you want to be without ever being told.
\\
\midrule

Derek didn't feel like he was going to try it. He took a swig of wine, stopped to look at the casket, pulled a pistol from the depths of his jacket, and shot himself in the head.

Over the past week, Derek had been learning how to say goodby to Jennifer. She was an avid drinker, if he hadn't already been about to drop her off at the bar, she would have never stopped talking. Derek understood that it was part of his life now, but he didn't want her taking the brunt of his life for something he didn't understand.

Derek knocked over a glass of coke as he heard the small women's voice whisper to him. ``I see your coat is getting in shape."

Derek had been worried about getting up to Jennifer. She always appeared in front of him when he talked, so he took a few steps to give him a room where he could whisper into her ear as she silently laughed with her mother. Derek had walked away from the bar, and headed to the bedroom of his father.

He sat down on the dresser, grabbed the phone, and dialed the number that was listed in the phonebook. He pulled up his phone to record his message and played it with the button.

``Hello?"

``I'm sorry I called to speak to you, but you've been shot dead. Your loved ones have been informed that your funeral will be in 20 minutes."

The tears were flowing, faster now. ``Why? Why didn't they do this?"

The last words Derek heard before he lost control of the gun were, ``I couldn't do it. I told them I wasn't happy, that it was too hard. I didn't mean to kill them, but they kept on telling me I wasn't a good man. And now I'm going to die again."

A tear dripped from the corner of his left eye and splattered over the glass of coke that was cradling his head. He didn't know if she was holding him back or trying to make him more crazy, but it was over and he was falling apart.

He began sobbing.
\\
\midrule
\textbf{Prompt:} North Korea's space program has been sabotaged. They are preparing for a final push for a country they
can not reach.
\\
\midrule
They've already prepared for the worst.

We all watched as the North Korean missile interceptor went off course, heading directly towards the Northern Sea. No one seemed to be very concerned about their fate, everyone was simply curious. But when North Korea launched their missiles in four straight days, things started to go horribly wrong.

The ``first in a long line of spaceships" North Korean propaganda portrayed, had been mysteriously appearing in the sky over the course of the few hours North Korea had taken to launch their first space craft. No one suspected anything, and nobody
really cared. The entire military was horrified at what they were seeing.

That's when the horrible explosion they had seen coming at the end of their first space craft caused North Korea to launch
it's first space flight.

Many initially feared it was an act of war. That it was coming for them with what seemed like precision, purpose. Instead,
it was merely a sick piece of manufactured shit, a broken economy. They saw it coming for the very first time, because their weapons were tested, then they just killed each other over the course of the next 48 hours.

The end result was far more interesting to everyone. China, Russia, and Japan all surrendered, saying that they didn't want to risk another war in the middle of the Pacific. Just as they all decided it was a self-defense measure, and that was it.

Korea was prepared for what they were about to see. But what they saw was much, much worse than the initial reports. It was
a massive wall of nuclear powered missiles. Before they knew it, the entire world knew what was about to happen, and would happen for all of humanity.
\\
\bottomrule
\end{tabularx}
\end{center}
\caption{Samples generated by GPT2 $\rightarrow$ BC $\rightarrow$ WP + SWAG + SYNTH.}
\end{table*}

\begin{table*}
\begin{center}
\begin{tabularx}{\textwidth}{X}
\toprule
\textbf{Prompt:} The scientists have discovered something terrible.
\\
\midrule
Robert Burton looked up at the scientists, his stomach churned. The lead researcher of the institute, Sir Peter Hetty, was simply simply not interested. ``This is absurd, but they have to make a case."

``To prove to the world that the M'ledvian radiation has been `neutralized' rather than being harmless," replied the human researcher, Steve Buckley. The room fell quiet once again. ``So then, we can set up a containment field. That way no more would we be irradiated? Why would you want to take any precautions with M'ledvian radiation?"

``With regard to this, Sir," repeated Robert, ``I simply can not accept that M'ledvian radiation has been found to be harmless."

``And what the *hell* would you do, give up on anything, give up your thesis to write about what would happen if a planet's
moon was dropped?" Buckley asked in a full tirade. ``What sort of miscalculation would that get us?"

``But we still can't just accept that there's no life on the moon. The black hole they created to completely incinerate us doesn't even have that great of a gravitational pull we are left with." Burton said, his voice rising.

``So we should be doing it, working to protect all life on this rock from us. Is that too much to ask, Jeremy?" Asked Buckley, spitting at the ground.

``Exactly. If we give up on doing it now, with the orbital research we did with the Mars Project, we would have no chance, just like we did before the bombs dropped." Buckley continued, sipping his tea.

``Alright, Buckley. Well done. We shall begin. " Burton smiled as he went back to his tea. ``May we get some lunch, please?"
\\
\bottomrule
\end{tabularx}
\end{center}
\caption{Samples generated by GPT2 $\rightarrow$ BC $\rightarrow$ WP + SWAG + SYNTH primed with the same prompts as \citet{DBLP:conf/acl/LewisDF18}.}
\end{table*}

\end{document}